\documentclass[journal, twocolumn]{IEEEtran}

\usepackage{amsmath,graphicx}
\usepackage{color}

\usepackage{lineno,hyperref}
\usepackage{amssymb}
\usepackage{amsmath,bm}
\usepackage{booktabs}
\usepackage{color}
\usepackage{amsmath,graphicx}
\usepackage{color}
\usepackage{booktabs}
\usepackage{color}
\usepackage{multirow}
\usepackage{makecell}

\usepackage[ruled,linesnumbered]{algorithm2e}
\usepackage{amsmath,graphicx}
\usepackage{color}

\usepackage{lineno,hyperref}
\usepackage{amssymb}
\usepackage{amsmath,bm}
\usepackage{booktabs}
\usepackage{color}

\usepackage{multirow}
\usepackage{makecell}
\usepackage{bm}

\newcommand{\tabincell}[2]{\begin{tabular}{@{}#1@{}}#2\end{tabular}} 
\usepackage{cite}

\ifCLASSINFOpdf
\else
\fi

\begin{document}
	
	\title{Hi-Net: Hybrid-fusion Network for Multi-modal MR Image Synthesis}
	
	\author{Tao Zhou, Huazhu Fu, Geng Chen, Jianbing Shen, \IEEEmembership{Senior Member, IEEE}, and Ling Shao.
		\thanks{Corresponding author: \emph{Huazhu Fu, Jianbing Shen}.}
		\thanks{T. Zhou, H. Fu, G. Chen, J. Shen, and L. Shao are with Inception Institute of Artificial Intelligence, Abu Dhabi, UAE.  
			(e-mails:\{tao.zhou, huazhu.fu, geng.chen, jianbing.shen, ling.shao\}@inceptioniai.org).}}

	\maketitle

	\begin{abstract}
		
		Magnetic resonance imaging (MRI) is a widely used neuroimaging technique that can provide images of different contrasts (\emph{i.e.}, modalities). Fusing this multi-modal data has proven particularly effective for boosting model performance in many tasks. However, due to poor data quality and frequent patient dropout, collecting all modalities for every patient remains a challenge. Medical image synthesis has been proposed as an effective solution to this, where any missing modalities are synthesized from the existing ones. In this paper, we propose a novel Hybrid-fusion Network (Hi-Net) for multi-modal MR image synthesis, which learns a mapping from multi-modal source images (\emph{i.e.}, existing modalities) to target images (\emph{i.e.}, missing modalities). In our Hi-Net, a modality-specific network is utilized to learn representations for each individual modality, and a fusion network is employed to learn the common latent representation of multi-modal data. Then, a multi-modal synthesis network is designed to densely combine the latent representation with hierarchical features from each modality, acting as a generator to synthesize the target images. Moreover, a layer-wise multi-modal fusion strategy is presented to effectively exploit the correlations among multiple modalities, in which a Mixed Fusion Block (MFB) is proposed to adaptively weight different fusion strategies (\emph{i.e.}, element-wise summation, product, and maximization). Extensive experiments demonstrate that the proposed model outperforms other state-of-the-art medical image synthesis methods. 
		
	\end{abstract}
	

	\section{Introduction}

	\IEEEPARstart{M}{edical} imaging plays an important role in a variety of clinical applications. In particular,  magnetic resonance imaging (MRI) is a versatile and noninvasive imaging technique extensively used in disease diagnosis, segmentation and other tasks \cite{zhou2019deep,zhang2018translating,zhou2019inter,ma2018concatenated,fan2019adversarial}. MRI comes in several modalities, such as $T_1$-weighted ($T_1$), $T_1$-with-contrast-enhanced ($T_{1c}$), $T_2$-weighted ($T_2$), and $T_2$-fluid-attenuated inversion recovery ($Flair$). Because each modality captures specific characteristics of the underlying anatomical information, combining multiple complementary modalities can provide a highly comprehensive set of data. Thus, several studies have focused on integrating the strengths of multiple modalities by exploring their rich information and discovering the underlying correlations among them, as a means of improving various medical tasks \cite{liu2018digital,zhu2017novel,shen2019brain,zhu2016subspace,tong2017multi}.

	In clinical application, however, acquiring multiple MR imaging modalities is often challenging for a variety of reasons, such as scan cost, limited availability of scanning time, and safety considerations. This inevitably results in incomplete datasets and adversely affects the quality of diagnosis and treatment in clinic analysis. Since most existing methods are not designed to cope with missing modalities, they often fail under these conditions. One solution is to simply discard samples that have one or more missing modalities and perform tasks using the remaining samples with complete multi-modal data. However, this simple strategy disregards lots of useful information contained in the discarded samples and also escalates the small-sample-size issue. To overcome this, cross-modal medical image synthesis has gained widespread popularity, as it enables missing modalities to be produced artificially without requiring the actual scans. Currently, many learn-based synthesis methods have been proposed and obtained promising performance  \cite{Jog2013,Torrado2016,huang2017cross,dong2014learning,Nie2017,Wang2018tmi}.

	Currently, a large portion of medical image synthesis methods work on single-modality data \cite{huang2017cross,Huang2017,Nie2017,Wolterink2017,Wang2018tmi,dar2019image}. However, because multiple modalities are often used in medical applications, and based on the principle ``\emph{more modalities provide more information}", several studies have begun investigating multi-modal data synthesis. For example, Chartsias \emph{et al.} proposed a multi-input multi-output method for MRI synthesis. Olut \emph{et al.}  \cite{olut2018generative} proposed a synthesis model for generating MR angiography sequences using the available $T_1$ and $T_2$ images. Yang \emph{et al.}  \cite{yang2019bi} proposed a bi-modal medical image synthesis method based on a sequential GAN model and semi-supervised learning. However, there still remains a general dearth of methods that use multi-modal data as input to synthesize medical images. To achieve a multi-modal synthesis, one critical challenge is effectively fusing the various inputs. One fusion strategy is to learn a shared representation \cite{yi2015shared,chen2015generalized,peng2016cross}.  For example, Ngiam \emph{et al.} \cite{ngiam2011multimodal} used a bi-modal deep autoencoder to fuse auditory and visual data, employing shared representation learning to then reconstruct the two types of inputs. Shared representation learning has proven particularly effective for exploiting the correlations among multi-modal data. However, while exploiting these correlations is important, preserving modality-specific properties is also essential for the multi-modal learning task, thus it is challenging to automatically balance the two aspects.

	To this end, we propose a novel Hybrid-fusion Network (Hi-Net) for multi-modal MR image synthesis, which synthesizes the target (or missing) modality images by fusing the existing ones. Specifically, our model first learns a modality-specific network to capture information from each individual modality. This network is formed as an autoencoder to effectively learn the high-level feature representations. Then, a fusion network is proposed to exploit the correlations among multiple modalities. In addition, we also propose a layer-wise multi-modal fusion strategy that can effectively exploit the correlations among different feature layers. Furthermore, an MFB is presented to adaptively weight different fusion strategies (\emph{i.e.}, element-wise summation, product, and maximization). Finally, our Hi-Net combines the modality-specific networks and fusion network to learn a latent representation for the various modalities, in which it is used to generate the target images. The effectiveness of the proposed synthesis method is validated by comparing it with various existing state-of-the-art methods \footnote{Code is publicly available at: https://github.com/taozh2017/HiNet}.
	
	The main contributions of this paper are listed as follows. 
	
	\begin{itemize}{\setlength{\parsep}{-0.25ex}}
		
		\item Different from most existing single-modality synthesis approaches, we present a novel medical image synthesis framework that uses multiple modalities to synthesize target-modality images.
		
		\item Our model captures individual modality features through the modality-specific network, as well as exploits the correlations among multiple modalities using a layer-wise multi-modal fusion strategy to effectively integrate multi-modal multi-level representations.
		
		\item A novel MFB module is proposed to adaptively weight the different fusion strategies, effectively improving the fusion performance. 
		
		
	\end{itemize}
	
	
	The rest of this paper is organized as follows. We introduce some related works in Section~\ref{Materials}. Then, we describe the framework of our proposed Hi-Net for medical image synthesis in Section~\ref{Methodology}. We further present the experimental settings, experimental results, and discussion in Section~\ref{Experiments}. Finally, we conclude the paper in Section~\ref{Conclusion}.

	\section{Related Works}
	\label{Materials}
	
	
	We review some related works on the cross-modal synthesis, medical image synthesis, and multi-modal learning below.
	
	\textbf{Cross-modal synthesis}. Recently, GAN-based cross-modal synthesis methods have attracted significant interest and achieved great success \cite{Goodfellow,pix2pix,cGAN,cycleGAN}. The key idea is to conduct continuous adversarial learning between a generator and a discriminator, where the generator tries to produce images that are as realistic as possible, while the discriminator tries to distinguish the generated images from real ones. For example, Pix2pix \cite{pix2pix} focuses on pixel-to-pixel image synthesis based on paired data, reinforcing the pixel-to-pixel similarity between the real and the synthesized images. The conditional GAN \cite{cGAN} was proposed to learn a similar translation mapping under a conditional framework to capture structural information. Besides, CycleGAN \cite{cycleGAN} was proposed to generalize the conditional GAN and can be applied to unpaired data.

	\begin{figure*}
		\begin{center}
			\includegraphics[width=0.89\textwidth]{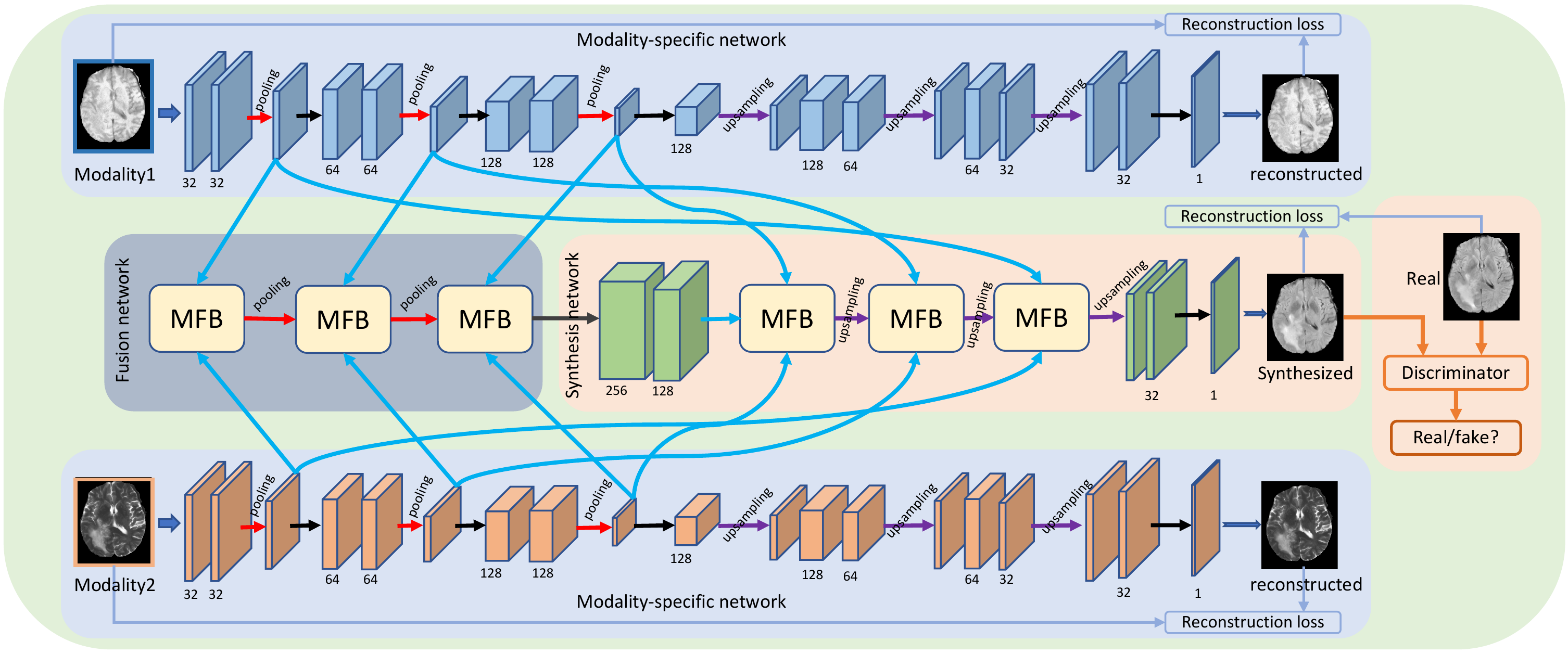}\vspace{-0.2cm}
			\caption{Framework of the proposed Hi-Net for multi-modal medical image synthesis. Our Hi-Net includes three main components: the modality-specific network, multi-modal fusion network, and multi-modal synthesis network. The modality-specific network is used to learn modality-specific properties, while the multi-modal fusion network aims to learn the correlations among multiple modalities. The multi-modal synthesis network consists of a generator and a discriminator, where the generator network synthesizes the target images, and the discriminator aims to distinguish the synthesized and real images.
			}
			\label{fig1}
		\end{center}\vspace{-0.35cm}
	\end{figure*}
	
	\textbf{Medical image synthesis}. Several machine learning-based synthesis methods have been developed. Traditional synthesis methods are often regarded as \emph{patch-based} regression tasks \cite{Jog2013,Torrado2016}, which take a patch of an image or volume from one modality to predict the intensity of a corresponding patch in a target image. Next, a regression forest was utilized to regress target-modality patches from these given-modality patches. In addition to the patch-based regression models, there has also been rapid development in \emph{sparse representation} for medical image synthesis \cite{huang2017cross,Huang2017}. In \cite{Huang2017}, Huang \emph{et al.}  proposed a weakly supervised joint convolutional sparse coding to simultaneously solve the problems of super-resolution (SR) and cross-modality image synthesis. Another popular type of medical image synthesis method is the \emph{atlas-based} model \cite{Miller1993}. These methods \cite{Roy2013,Burgos2014} adopted the paired image atlases from the source- and target-modalities to calculate the atlas-to-image transformation in the source-modality, which is then applied for synthesizing target-modality-like images from their corresponding target-modality atlases. More recently, deep learning has been widely applied in medical image analysis, achieving promising results. For the image synthesis task, Dong \emph{et al.} \cite{dong2014learning} proposed an end-to-end mapping between low/high-resolution images using convolutional neural networks (CNNs). Li \emph{et al.} \cite{li2014deep} used a deep learning model to estimate the missing positron emission tomography (PET) data from the corresponding MRI data. In addition, GAN-based methods have also achieved promising results in synthesizing various types of medical images, such as CT images \cite{Nie2017}, retinal images \cite{costa2017end,costa2017towards}, MR images \cite{dar2019image}, ultrasound images \cite{tom2018simulating,yi2018generative} and so on. For example, Nie \emph{et al.} \cite{Nie2017} utilized MR images to synthesize computed tomography (CT) images with a context-aware GAN model. Wolterink \emph{et al.} \cite{Wolterink2017} utilized GAN to transform low-dose CT into routine-dose CT images. Wang \emph{et al.} \cite{Wang2018tmi} also demonstrated promising results when using a GAN to estimate high-dose PET images from low-dose ones.  The list of GAN-based methods proposed for medical synthesis is extensive \cite{Yutmi2019,guibas2017synthetic,dar2019image,zhang2018translating,Zhang2019}.

	\textbf{Multi-modal learning}. Many real-world applications involve multi-modal learning, since data usually can often be obtained from multiple modalities. Due to the effectiveness of exploring the complementarity among multiple modalities, multi-modal learning has attracted increased attention recently. One popular strategy is to find a new space of common components that can be more robust than any of the input features from different modalities. For example, canonical correlation analysis (CCA) \cite{hardoon2004canonical} projects the features of each modality to a lower-dimensional subspace. Multiple kernel learning (MKL) \cite{lin2010multiple} utilizes a set of predefined kernels from multi-view data and integrates these modalities using the optimized weights. In addition, there are several works that have applied deep networks to multi-modal learning. For example, Zhou \emph{et al.} \cite{zhou2019effective} presented a three-stage deep feature learning framework to detect disease status via fusing MRI, PET, and single-nucleotide polymorphism (SNP) data. Nie \emph{et al.} \cite{nie20163d} proposed a 3D deep learning model to predict the overall survival time for brain gliomas patients by fusing $T_1$ MRI, functional (fMRI) and diffusion tensor imaging (DTI) data. Wang \emph{et al.} \cite{wang2015large} proposed a multi-modal deep learning framework for RGB-D object recognition, which simultaneously learns transformation matrices for two modalities and a maximal cross-modality correlation criterion. Hou \emph{et al.} \cite{hou2019deep} proposed a high-order polynomial multilinear pooling block to achieve multi-modal feature fusion, in which a hierarchical polynomial fusion network can flexibly fuse the mixed features across both time and modality domains. Some of the above methods \cite{zhou2019effective,nie20163d} focus on fusing the features from multi-modalities in high-level layers, while our model proposes a hybrid-fusion network to integrate multi-modal multi-level representations.

	\section{Methodology}
	\label{Methodology}
	
	In this section, we provide details for the proposed Hi-Net, which is comprised of three main components: the modality-specific network, multi-modal fusion network, and multi-modal synthesis network (consisting of a generator and a discriminator).

	\subsection{Modality-specific Network} 
	
	In multi-modal learning, complementary information and correlations from multiple modalities are expected to boost the learning performance. Thus, it is critical to exploit the underlying correlations among multiple modalities, while also capturing the modality-specific information to preserve their properties. To achieve this goal, we first construct a modality-specific network for each individual modality (\emph{e.g.}, $x_i$), as shown in Fig.~\ref{fig1}. Thus, the high-level feature representation $h_{x_i}$ for the $i$-th modality can be represented as $h_{x_i}=f_{\Theta_{i}^{en}}(x_i)$, where $\Theta_{i}^{en}$ denotes the network parameters. To learn a meaningful and effective high-level representation, we adopt an autoencoder-like structure to reconstruct the original image using the learned high-level representation. To do so, we have the following reconstruction loss function:
	\begin{eqnarray}
	\begin{split}
	\mathcal{L}^{R}=\sum \nolimits_{i}\|x_i-\hat{x_i}\|_1,
	\end{split}
	\label{eq2-001}
	\end{eqnarray}
	\noindent where $\hat{x_i}=f_{\Theta_{i}^{de}}(x_i)$ denotes the reconstructed image of $x_i$, and $\Theta_{i}^{de}$ denotes the corresponding network parameters. Besides,  we also use the $\ell_1$-norm to measure the difference between the original and reconstructed images. It is worth noting that the reconstruction loss provides side-output supervision to guarantee that the modality-specific network learns a discriminative representation for each individual modality.

	A detailed illustration of the modality-specific network can be found in Fig.~\ref{fig1}. In each convolutional layer, we use a $3\times{3}$ filter with stride 1 and padding 1. Besides, we also introduce batch normalization after each convolutional layer. Specifically, each batch is normalized during the training procedure using its mean and standard deviation, and then global statistics are computed from these values. After the batch normalization, the activation functions \emph{LeakyReLu} and \emph{ReLu} are used in the encoder and decoder, respectively. The pooling and upsampling layers use $2\times{2}$ filters.

	\subsection{Multi-modal Fusion Network} 
	
	Most existing multi-modal learning approaches employ one of two multi-modal feature fusion strategies, \emph{i.e.}, early fusion or late fusion. Early fusion directly stacks all raw data and then feeds into a single deep network, while late fusion first extracts high-level features from each modality and then combines them using a concatenation layer. To effectively exploit the correlations between multi-level representations from different layers (\emph{e.g.}, shallow layers and high-level layers) and reduce the diversity between different modalities, we propose a layer-wise fusion network. Moreover, an MFB module is also proposed to adaptively weight different inputs from various modalities. As shown in Fig. \ref{fig1}, the feature representations from the first pooling layer of each modality-specific network are fed into an MFB block, then the output of this front MFB module is input into the next MFB module with the feature representations of the second pooling layer of the modality-specific network. Thus, we have three MFB modules in the fusion network, as shown in Fig. \ref{fig1}. It is worth noting that the layer-wise fusion is independent of the modality-specific network, thus it does not disturb the modality-specific structure and only learns the underlying correlations among the modalities. Besides, the proposed layer-wise fusion is conducted in different layers, thus our model can exploit the correlations among multiple modalities using low-level as well as high-level features.

\begin{figure}
		\begin{center}
			\includegraphics[width=0.45\textwidth]{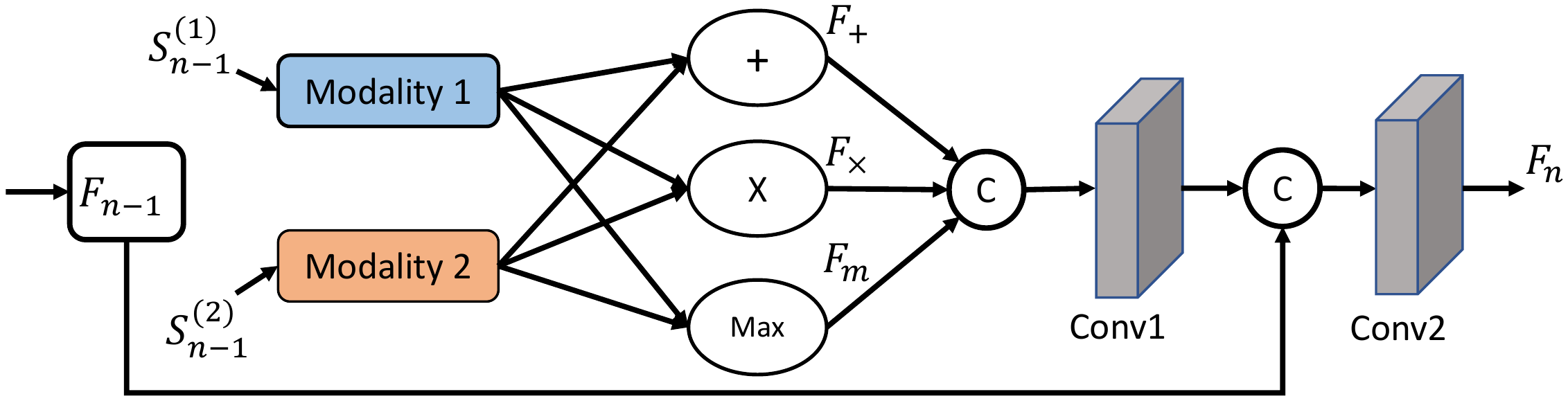}\vspace{-0.1cm}
			\caption{Structure of the MFB module (where ``+", ``$\times$", ``Max", and ``c" denote the element-wise summation, element-wise product, element-wise maximization and concatenation operations, respectively).}
			\label{fig2}
		\end{center}\vspace{-0.35cm}
	\end{figure}

	Fig.~\ref{fig2} provides an illustration of the MFB module, where an adaptive weight network is designed to fuse the feature representations from multiple modalities. In the multi-modal fusion task, popular strategies include element-wise summation, element-wise product and element-wise maximization. However, it is not clear which is best for different tasks. Thus, to benefit from the advantages of each strategy, we simultaneously employ all three fusion strategies and then concatenate them. Then, a convolutional layer is added to adaptively weight the three fusions. As shown in Fig.~\ref{fig2}, we obtain the output $S_{n-1}^{(i)}\in\mathbb{R}^{C\times{W}\times{H}}$ for the $(n-1)$-th pooling layer of the $i$-th modality, where $C$ is the number of feature channels, and $W$ and $H$ denote the width and height of the feature maps, respectively. Then, we apply the three fusion strategies to the inputs $S_{n-1}^{(i)} (i=1,2)$ to obtain
	\begin{equation}
	\left\{
	\begin{aligned}
	&F_{+}=S_{n-1}^{(1)}+ S_{n-1}^{(2)};\\
	&F_{\times}=S_{n-1}^{(1)}\times S_{n-1}^{(2)};\\
	&F_{m}=\textup{Max}(S_{n-1}^{(1)}, S_{n-1}^{(2)}),
	\end{aligned}
	\right.
	\label{eq91}
	\end{equation}
	\noindent where ``$+$", ``$\times$" and ``\textup{Max}" denote element-wise summation, element-wise product and element-wise maximization operations, respectively. Then, we combine them as $F_{concat}=[F_{+};F_{\times};F_{m}]\in\mathbb{R}^{3C\times{W}\times{H}}$. $F_{concat}$ is then fed into the first convolutional layer (\emph{i.e.,} Conv1 in Fig.~\ref{fig2}). The output of this layer is concatenated with the previous output $F_{n-1}$ of the $(n-1)$-th MFB module, and fed into the second convolutional layer (\emph{i.e.,} Conv2 in Fig.~\ref{fig2}). Finally, we obtain the output $F_n$ of the $n$-th MFB module. Note that, when $n=1$, there is no previous output $F_{n-1}$, so we simply feed the output of the first convoulutional layer into the \emph{Conv2} layer.  It is also worth noting that the fusion method allows the MFB module to adaptively weight the different feature representations from multiple modalities, which benefits from the above three fusion strategies.
	
	For the fusion network, the size of all filters is $3\times{3}$, and the numbers of filters are $32$ and $64$, $64$ and $128$, and $128$ and $128$ for the three MFB modules, respectively.  Batch normalization is conducted after each convolutional layer using the \emph{ReLu} activation function.

	\subsection{Multi-modal Synthesis Network}
	
	Once the multi-modal latent representation $F_n$ (\emph{i.e.}, the output of the last MFB module in the multi-modal fusion network) has been obtained, we can use it to synthesize a target-modality image via a GAN model. Similar to the pixel-to-pixel synthesis method \cite{pix2pix}, our generator $G^{'}$ tries to generate an image $G^{'}(F_n)$ from the input $F_n$, while the discriminator $D$ tries to distinguish the generated image $G^{'}(F_n)$ from the real image $y$. Accordingly, the objective function of the generator can be formulated as:
	\begin{eqnarray}
	\begin{split}
	\mathcal{L}^{G}=~\mathbf{E}_{F_n\sim{p_{data}}}&[\log(1-D(F_n,(G^{'}(F_n))))]\\&+\lambda_1\mathbf{E}_{F_n,y}[\|y-G^{'}(F_n)\|_1],
	\end{split}
	\label{eq1-002}
	\end{eqnarray}
	\noindent where $\lambda_1$ is a nonnegative trade-off parameter.  The generator $G^{'}$ tries to generate a realistic image that misleads $D$ in the first term of Eq. (\ref{eq1-002}), and an $\ell_1$-norm is used to measure the difference between the generated image and the corresponding real image in the second term. Because we  integrate multi-modal learning and image synthesis into a unified framework, the generator can be reformulated as:
	\begin{eqnarray}
	\begin{split}
	\mathcal{L}^{G}=~\mathbf{E}_{x_1,x_2\sim{p_{data}}}&[\log(1-D(x_1,x_2,(G(x_1,x_2))))]\\&+\lambda_1\mathbf{E}_{x_1,x_2,y}[\|y-G(x_1,x_2)\|_1].
	\end{split}
	\label{eq1-003}
	\end{eqnarray}

	Additionally, the objective function of the discriminator $D$ can be formulated as:
	\begin{eqnarray}
	\begin{split}
	\mathcal{L}^{D}=~-&\mathbf{E}_{y\sim p_{data}}[\log D(y)]\\
	&-\mathbf{E}_{x_1,x_2\sim p_{data}}[\log(1-D(G(x_1,x_2)))].
	\end{split}
	\label{eq1-004}
	\end{eqnarray}
	
	Finally, an end-to-end multi-modal synthesis framework can be formulated with the following objective:
	\begin{eqnarray}
	\begin{split}
	\mathcal{L}=\mathcal{L}^G+\mathcal{L}^{D}+\lambda_2\mathcal{L}^{R},
	\end{split}
	\label{eq2-002}
	\end{eqnarray}
	\noindent where $\lambda_2$ is a trade-off parameter.
	
	The detailed architecture of the generator is shown in Fig.~\ref{fig1}. Specifically, we first feed the latent representation into two convolutional layers with $256$ and $128$ filters of size $3\times{3}$, respectively, and then the output is further fed into three MFB modules. Note that we also use the MFB modules to fuse the latent representation and the feature representations from the encoding layers of each modality-specific network using a skip connection. Then, we feed the output of the last MFB module into an upsampling layer and two convolutional layers (with a filter size of $3\times{3}$ and number of filters $32$ and $1$, respectively). Batch normalization is also conducted after each convolutional layer, using a \emph{ReLu} activation function. 
	

	\begin{figure*}
		\begin{center}
			\includegraphics[width=0.9\textwidth]{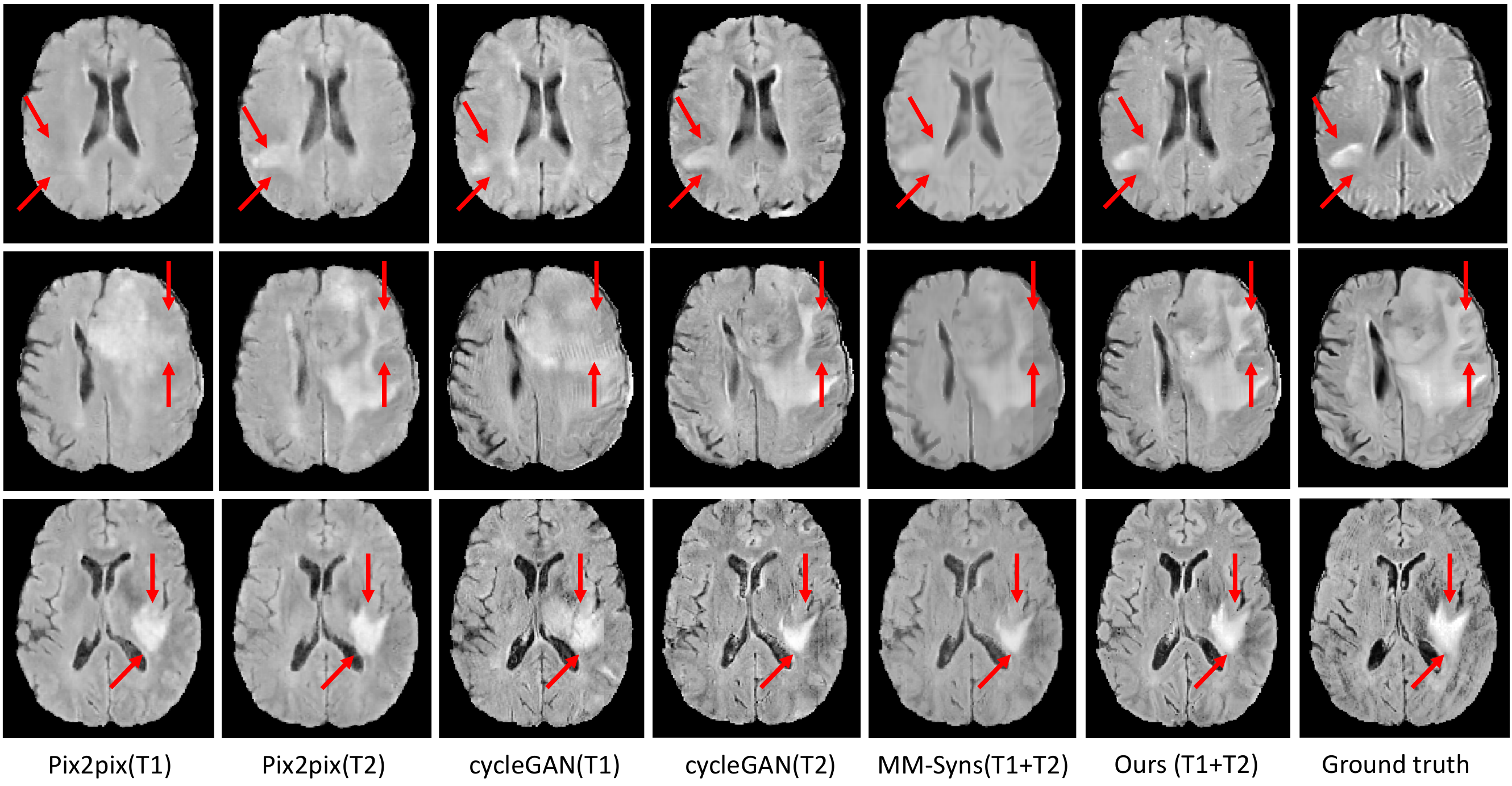}\vspace{-0.15cm}
			\caption{Qualitative comparison between the proposed synthesis method and other state-of-the-art methods for the  $Flair$ synthesis task on the BraTs2018 dataset}. Note that our method and MM-Syns use both $T_1$ and $T_2$ modalities, while the other methods use any one of the two. 
			\label{fig4}
		\end{center}\vspace{-0.35cm}
	\end{figure*}

	\begin{figure*}
		\begin{center}
			\includegraphics[width=0.9\textwidth]{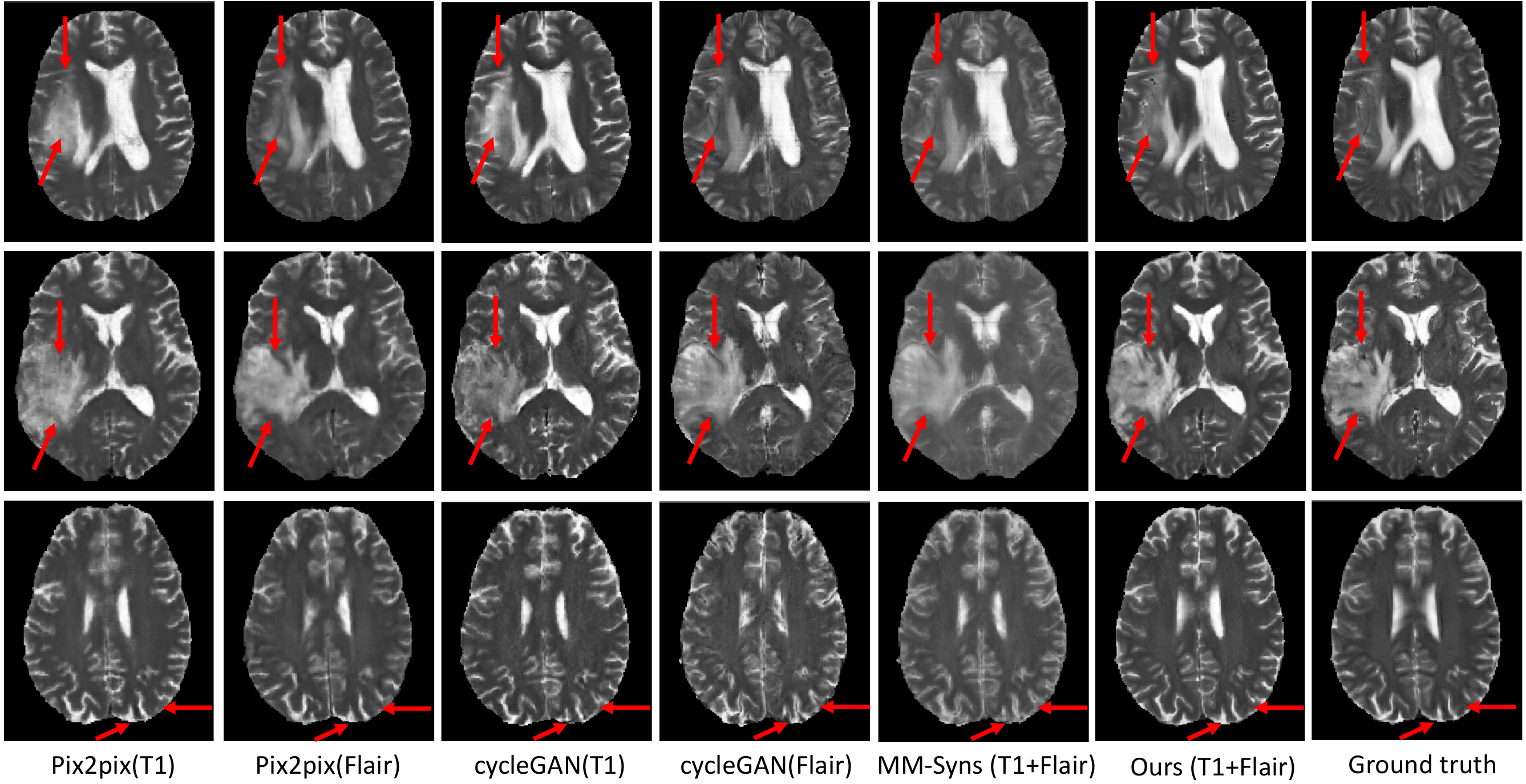}\vspace{-0.3cm}
			\caption{Qualitative comparison between the proposed synthesis method and other state-of-the-art synthesis methods for the $T_2$ synthesis task on the BraTs2018 dataset. Note that our method and MM-Syns use both $T_1$ and $Flair$ modalities, while the other methods use any one of the two. }
			\label{fig5}
		\end{center}\vspace{-0.35cm}
	\end{figure*}

	Additionally, the discriminator takes either a real target modality image or a synthesized one as input, and aims to determine whether  or not it is real. The input of the discriminator is a 2D image (\emph{e.g.,} $128\times 128$ in our experiments), which has the same size as the generator’s output. The architecture of the discriminator includes four convolutional layers, as it is defined as $\textup{Input}\rightarrow \textup{Conv}(3\times 3,32)\rightarrow \textup{BN}\rightarrow\textup{LeakyReLu}\rightarrow \textup{Conv}(3\times 3,64)\rightarrow \textup{BN}\rightarrow\textup{LeakyReLu}\rightarrow \textup{Conv}(3\times 3,128)\rightarrow \textup{BN}\rightarrow\textup{LeakyReLu}\rightarrow \textup{Conv}(3\times 3,256)\rightarrow \textup{BN}\rightarrow\textup{LeakyReLu}\rightarrow \textup{Conv}(3\times 3,1)\rightarrow \textup{Output}$, where BN denotes batch normalization. For the first four convolutional layers, we use filters with stride 2. Besides, all \emph{LeakyReLu} activations are with slope of 0.2.

	\section{Experiments and Results}
	\label{Experiments}
	
	In this section, we describe our experimental settings, including the dataset, comparison methods, evaluation metrics, and implementation details. We present comparison results, results for the ablation study, and some related discussion.

	\subsection{Dataset}
	To validate the effectiveness of our model, we use the multimodal brain tumor segmentation challenge 2018 (BraTs2018) dataset \cite{brast}. This dataset consists of 285 patients with multiple MR scans acquired from 19 different institutions and includes glioblastoma (GBM) and lower grade glioma (LGG) cohorts. The patient scans contain four modalities of co-registered MR volumes: $T_1$, $T_{1c}$, $T_2$, and $Flair$, where each modality volume is of size $240\times240\times155$. In this study, we use $T_1$, $T_2$ and $Flair$ images to verify the effectiveness of our proposed synthesis method. Our architecture uses 2D axial-plane slices of the volumes. For a 2D slice ($240\times240$),  we crop out an image of size $160\times 180$ from the center region. Besides, we randomly split the 285 subjects to 80\% for training and 20\% for testing. To increase the number of training samples, we split each cropped image ($160\times180$) into four overlapping patches of size $128\times128$, and the overlapped regions are averaged to form the final estimation.  For each volume, we linearly scale the original intensity values to $[-1,1]$.

	\subsection{Comparison Methods and Evaluation Metrics}
	
	To verify the effectiveness of the proposed synthesis method, we compare it with three state-of-the-art cross-modality synthesis methods, Pix2pix \cite{pix2pix}, cycleGAN \cite{cycleGAN} and MM-Syns \cite{MMsyn}. These methods can be summarized as follows:  1) \textbf{Pix2pix} \cite{pix2pix}. This method synthesizes a whole image by focusing on maintaining the pixel-wise intensity similarity; 2) \textbf{cycleGAN} \cite{cycleGAN}. This method uses a cycle consistency loss to enable training without the need for paired data. In our comparison, we use the paired data to synthesize medical images from one modality to another; and 3) \textbf{MM-Syns} \cite{MMsyn}. This method first learns a common representation for multi-modal data and then synthesizes an MR image slice-by-slice under the constraint of pixel-wise intensity difference. 
	

	

	

	To quantitatively evaluate the synthesis performance, three popular metrics \cite{Yutmi2019} are adopted in this study: 1) Peak Signal-to-Noise Ratio (PSNR). Given a ground-truth image $y(x)$ and a generated image $G(x)$, PSNR is defined as 
	$\textup{PSNR}=10\log_{10}\frac{\textup{max}^2\left(y(x),G(x)\right)}{\frac{1}{N}\sum\|y(x)-G(x)\|_2^2}$, where $N$ is the total number of voxels in each image, and $\textup{max}^2(y(x),G(x))$ is the maximal intensity value of the ground-truth image $y(x)$ and the generated image $G(x)$; 2) Normalized Mean Squared Error (NMSE). This can be defined as 	$\textup{NMSE}=\frac{y(x)-G(x)}{\|y(x)\|_2^2}$; and 3) Structural Similarity Index Measurement (SSIM). This is defined as: $\textup{SSIM}=\frac{(2\mu_{y(x)}\mu_{G(x)}+c_1)(2\sigma_{y(x)G(x)}+c_2)}{(\mu_{y(x)}^2+\mu_{G(x)}^2+c_1)(\mu_{y(x)}^2+\mu_{G(x)}^2+c_2)}$, where $\mu_{y(x)}$, $\mu_{G(x)}$, $\sigma{y(x)}$, and $\sigma_{G(x)}$ are the means and variances of image $y(x)$ and $G(x)$, and $\sigma_{y(x)G(x)}$ is the covariance of $y(x)$ and $G(x)$. The positive constants $c_1$ and $c_2$ are used to avoid a null denominator. Note that a higher PSNR value, lower NMSE value, and higher SSIM value indicate higher quality in the synthesized image. 
	
	
	
	
	
	
	
	
	\subsection{Implementation Details}
	
	All networks are trained using the Adam solver. We conduct 300 epochs to train the proposed model. The original learning rate is set to 0.0002 for the first 100 epochs and then linearly decays to 0 over the remaining epochs. During training, the trade-off parameters $\lambda_1$ and $\lambda_2$ are set to $100$ and $20$, respectively. The code is implemented using the PyTorch library.

	\renewcommand\arraystretch{1.0}
	\begin{table}[t]
		\setlength{\belowdisplayskip}{0pt}
		\setlength{\abovedisplayskip}{0pt}
		\setlength{\abovecaptionskip}{0pt}
		\centering
		\scriptsize
		\caption{\footnotesize Quantitative evaluation results of the synthesized $Flair$ images using $T_1$ and $T_2$ or their combination for different methods on the BraTs2018 dataset (mean$\pm$standard deviation). Bold indicates the best result.}
		\begin{tabular}{p{3.3cm}|p{1.3cm}p{1.3cm}p{1.3cm}}  
			\toprule 
			\textbf{Methods}                                 & PSNR $\uparrow$    &NMSE $\downarrow	$        & SSIM $\uparrow$\\
			\toprule
			Pix2pix ($T_1\rightarrow Flair$)                &22.93$\pm$1.033        &0.0412$\pm$0.009       &0.8726$\pm$0.027\\
			Pix2pix ($T_2\rightarrow Flair$)                &22.86$\pm$0.918        &0.0419$\pm$0.008      &0.8694$\pm$0.029\\
			cycleGAN ($T_1\rightarrow Flair$)            &22.65$\pm$1.398        &0.0459$\pm$0.014       &0.8352$\pm$0.040\\
			cycleGAN ($T_2\rightarrow Flair$)            &22.42$\pm$1.379        &0.0499$\pm$0.010       &0.8344$\pm$0.028\\
			MM-Syns ($T1+T2\rightarrow Flair$)        &23.15$\pm$1.101        &0.0358$\pm$0.009        &0.8622$\pm$0.029\\
			Ours  ($T_1+T_2\rightarrow Flair$)           &\textbf{25.05}$\pm$\textbf{1.325}        &\textbf{0.0258}$\pm$\textbf{0.009}      &\textbf{0.8909}$\pm$\textbf{0.030} \\
			
			\toprule
		\end{tabular}
		\label{tab01}
	\end{table}

	\begin{figure*}
		\begin{center}
			\includegraphics[width=0.9\textwidth]{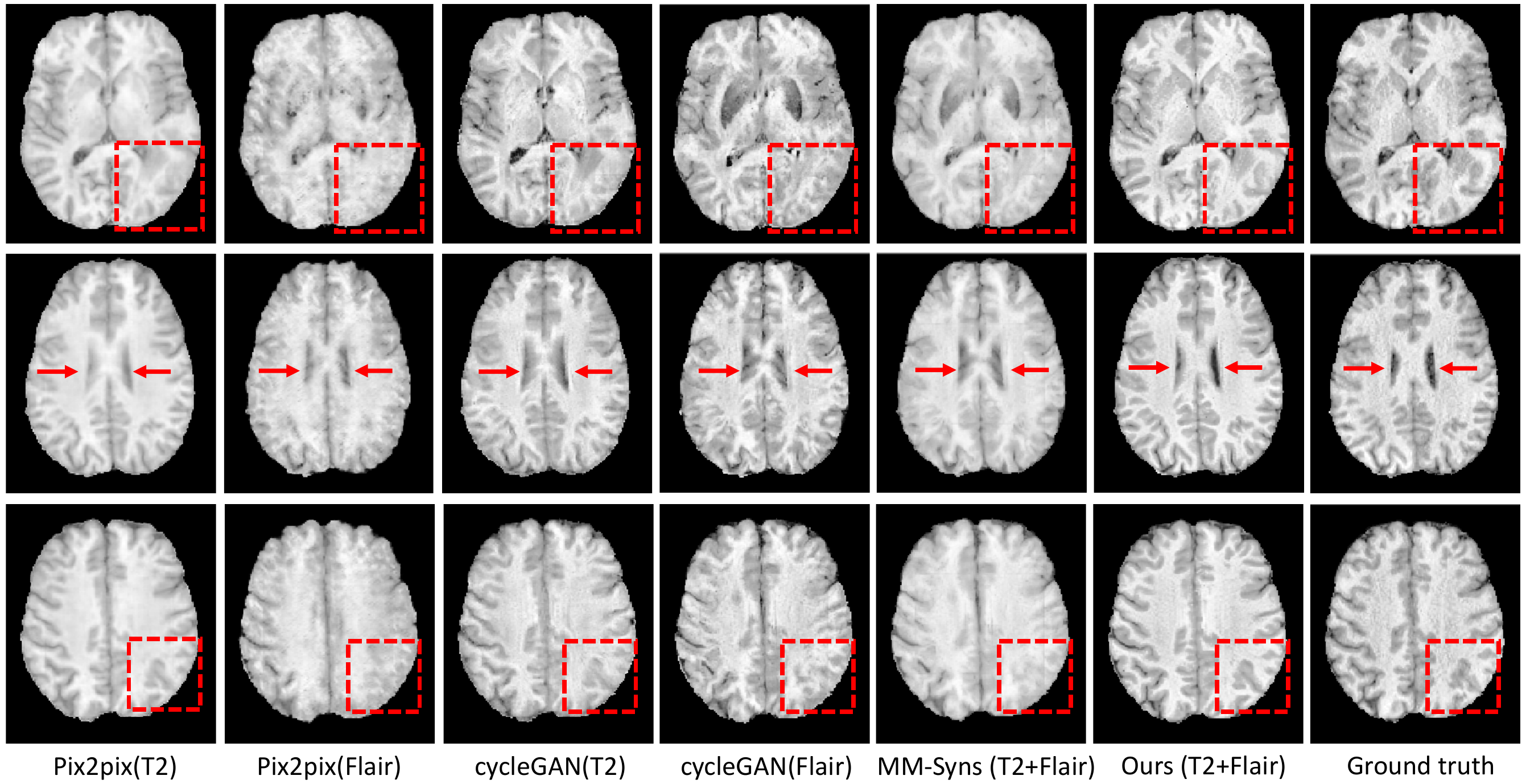}\vspace{-0.3cm}
			\caption{Qualitative comparison between the proposed synthesis method and other state-of-the-art synthesis methods for the $T_1$ synthesis task on the BraTs2018 dataset. Note that our method and MM-Syns use both $T_2$ and $Flair$ modalities, while other methods use any one of the two. }
			\label{fig6}
		\end{center}\vspace{-0.35cm}
	\end{figure*}
	
	\subsection{Results Comparison}
	
	We evaluate the proposed model for three tasks, \emph{i.e.,} on on BraTs2018 dataset, using $T_1$ and $T_2$ to synthesize the $Flair$ modality ($T_1+T_2\rightarrow{Flair}$), using $T_1$ and $Flair$ to synthesize the $T_2$ modality ($T_1+Flair\rightarrow{T_2}$), and using $T_2$ and $Flair$ to synthesize the $T_1$ modality ($T_2+Flair\rightarrow{T_1}$). We first evaluate the performance for synthesizing $Flair$ modality images. Table~\ref{tab01} shows the quantitative evaluation results for this task. From Table~\ref{tab01}, as it can be seen that our method outperforms all comparison methods in all three metrics (PSNR, NMSE, and SSIM). This suggests that our model can effectively fuse the complementary information from different modalities and that this is beneficial to the synthesis performance. Fig.~\ref{fig4} shows qualitative comparisons between the proposed synthesis method and other state-of-the-art methods in the $Flair$ synthesis task. As can be seen, our method achieves much better synthesis results. Specifically, our model can effectively synthesize tumor regions (as shown in the first row of Fig.~\ref{fig4}), while Pix2pix and cycleGAN either only synthesize some blurry tumor regions or fail to synthesize them. Similar observations can be found in the second and third rows of Fig.~\ref{fig4}. Overall, our method achieves much better synthesis results for the $Flair$ modality images, as demonstrated using both qualitative and quantitative measures.

	\renewcommand\arraystretch{1.0}
	\begin{table}[t]
		\setlength{\belowdisplayskip}{0pt}
		\setlength{\abovedisplayskip}{0pt}
		\setlength{\abovecaptionskip}{0pt}
		\centering
		\scriptsize
		\caption{\footnotesize Quantitative evaluation results of the synthesized $T_2$ images using $T_1$ and $Flair$ or their combination for different methods on the BraTs2018 dataset (mean$\pm$standard deviation). Bold indicates the best result.}
		\begin{tabular}{p{3.2cm}|p{1.3cm}p{1.3cm}p{1.3cm}}  
			\toprule 
			\textbf{Methods}                                   & PSNR $\uparrow$         &NMSE $\downarrow	$        & SSIM $\uparrow$ \\
			\toprule
			Pix2pix ($T_1\rightarrow T_2$)              &23.34$\pm$1.282          &0.0903$\pm$0.029      &0.8815$\pm$0.036\\
			Pix2pix ($Flair\rightarrow T_2$)             &23.44$\pm$1.267          &0.0882$\pm$0.027       &0.8746$\pm$0.034\\
			cycleGAN ($T_1\rightarrow T_2$)          &22.52$\pm$0.958          &0.1344$\pm$0.028       &0.8511$\pm$0.037\\
			cycleGAN ($Flair \rightarrow T_2$)        &22.38$\pm$0.874          &0.1259$\pm$0.033       &0.8328$\pm$0.028\\
			MM-Syns ($T_1+Flair\rightarrow T_2$)   &23.09$\pm$0.108          &0.1022$\pm$0.045       &0.8635$\pm$0.029\\
			Ours  ($T_1+Flair\rightarrow T_2$)         &\textbf{24.70}$\pm$\textbf{1.534}      &\textbf{0.0676}$\pm$\textbf{0.028}   &\textbf{0.9126}$\pm$\textbf{0.036}\\
			
			\toprule
		\end{tabular}
		\label{tab02}
	\end{table}

	For the second task (\emph{i.e.,} using $T_1$ and $Flair$ as inputs to synthesize $T_2$ images), Table~\ref{tab02} and Fig.~\ref{fig5} show the quantitative and qualitative comparison results for different methods, respectively. From Table~\ref{tab02} and Fig.~\ref{fig5}, as can be observed that our model outperforms other comparison synthesis methods. Additionally, in Fig.~\ref{fig5}, we can see that our method synthesizes better-quality target images compared to other methods. We have similar observations for the results shown in Table~\ref{tab03} and Fig.~\ref{fig6}, where our method obtains the best performance for the third task (\emph{i.e.,} using $T_2$ and $Flair$ as inputs to synthesize $T_1$ images) in terms of all evaluation metrics. It is also worth specifically noting that our method outperforms the MM-Syns method, which also uses two modalities to synthesize the target images, by using late fusion to learn the common representation and then reconstructs the target modality without using a GAN model. In contrast, our method presents a hybrid fusion network that can effectively explore complementary information from multiple modalities as well as exploit their correlations to improve synthesis performance. 
	
\begin{figure}
		\begin{center}
			\includegraphics[width=0.5\textwidth]{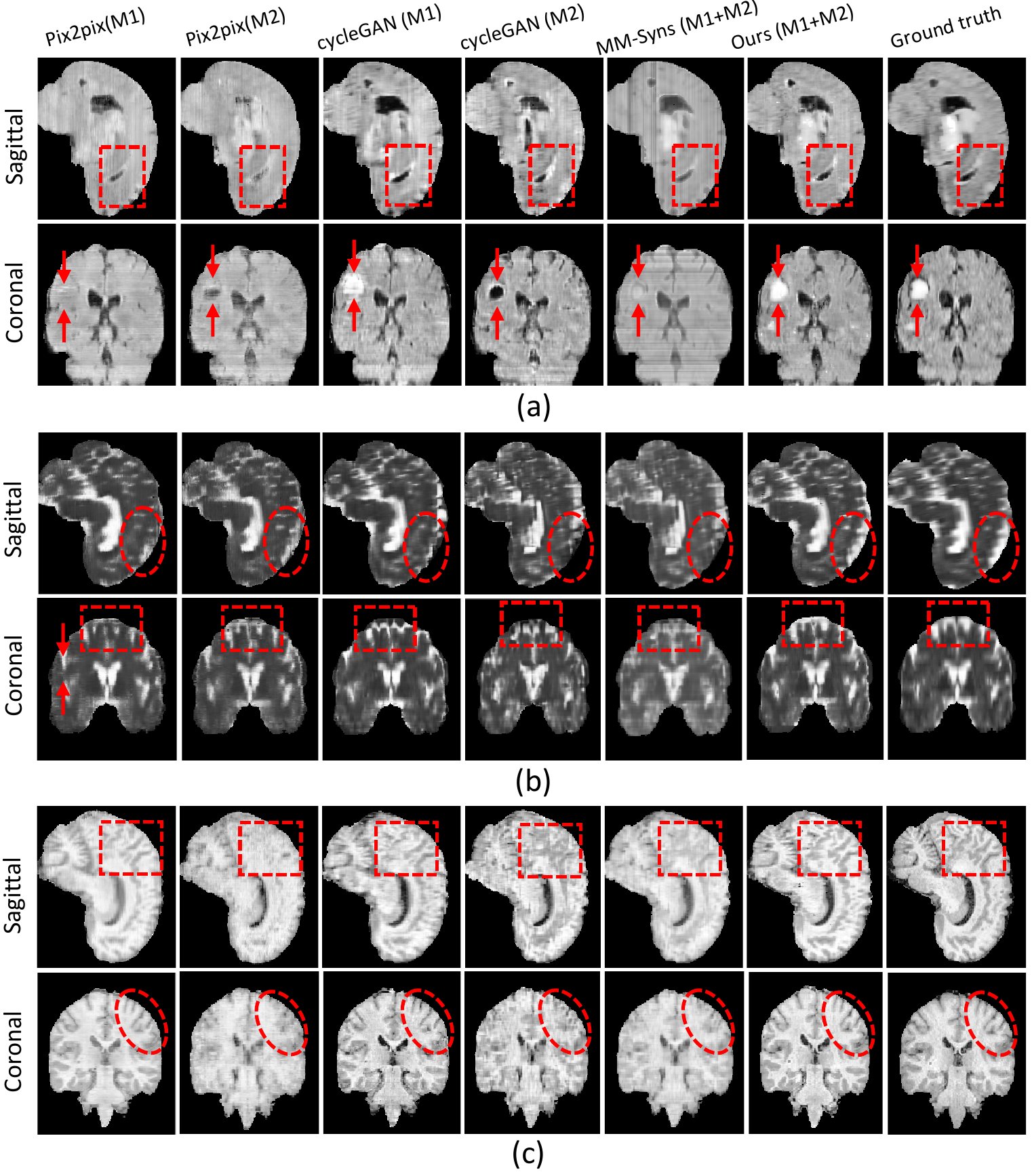}\vspace{-0.25cm}
			\caption{Qualitative results in sagittal and coronal planes using the proposed model and other state-of-the-art synthesis methods for three synthesis tasks {on the BraTS2018 dataset}. Note that in (a) M1: $T_1$, M2: $T_2$; (b) M1: $T_1$, M2: $Flair$; and (c) M1: $T_2$, M2: $Flair$.} 
			\label{fig7}
		\end{center}\vspace{-0.35cm}
	\end{figure}

	We have shown the comparison synthesis results in the axial plane, as slices are fed one-by-one into our synthesis network. To further validate the effectiveness of our model in the sagittal and coronal planes, we show the results for these in Fig.~\ref{fig7} within using different planes. From Fig.~\ref{fig7}, it can be seen that our method still performs better than other methods and is able to synthesize high-quality target images.

	\renewcommand\arraystretch{1.0}
	\begin{table}[t]
		\setlength{\belowdisplayskip}{0pt}
		\setlength{\abovedisplayskip}{0pt}
		\setlength{\abovecaptionskip}{0pt}
		\centering
		\scriptsize
		\caption{\footnotesize Quantitative evaluation results of the synthesized $T_1$ images using $T_2$ and $Flair$ or their combination for different methods on the BraTs2018 dataset (mean$\pm$standard deviation). Bold indicates the best result.}
		\begin{tabular}{p{3.2cm}|p{1.3cm}p{1.3cm}p{1.3cm}}  
			\toprule 
			\textbf{Methods}                                 & PSNR $\uparrow$    &NMSE $\downarrow	$        & SSIM $\uparrow$ \\
			\toprule
			Pix2pix ($T_2\rightarrow T_1$)                &23.74$\pm$1.140         &0.0248$\pm$0.007     &0.9036$\pm$0.030\\
			Pix2pix ($Flair\rightarrow T_1$)               &22.73$\pm$1.670         &0.0336$\pm$0.027     &0.8573$\pm$0.032\\
			cycleGAN ($T_2\rightarrow T_1$)            &23.36$\pm$1.045         &0.0269$\pm$0.007     &0.8862$\pm$0.031\\
			cycleGAN ($Flair\rightarrow T_1$)           &22.81$\pm$1.923          &0.0309$\pm$0.029     &0.8321$\pm$0.043\\
			MM-Syns ($T_2+Flair\rightarrow T_1$)    &22.94$\pm$0.821          &0.0329$\pm$0.011      &0.0833$\pm$0.027\\
			Ours  ($T_2+Flair\rightarrow T_1$)          &\textbf{25.23}$\pm$\textbf{1.387}          &\textbf{0.0180}$\pm$\textbf{0.007}      &\textbf{0.9145}$\pm$\textbf{0.033} \\
			
			\toprule
		\end{tabular}
		\label{tab03}
	\end{table}

\begin{figure*}
		\begin{center}
			\includegraphics[width=0.9\textwidth]{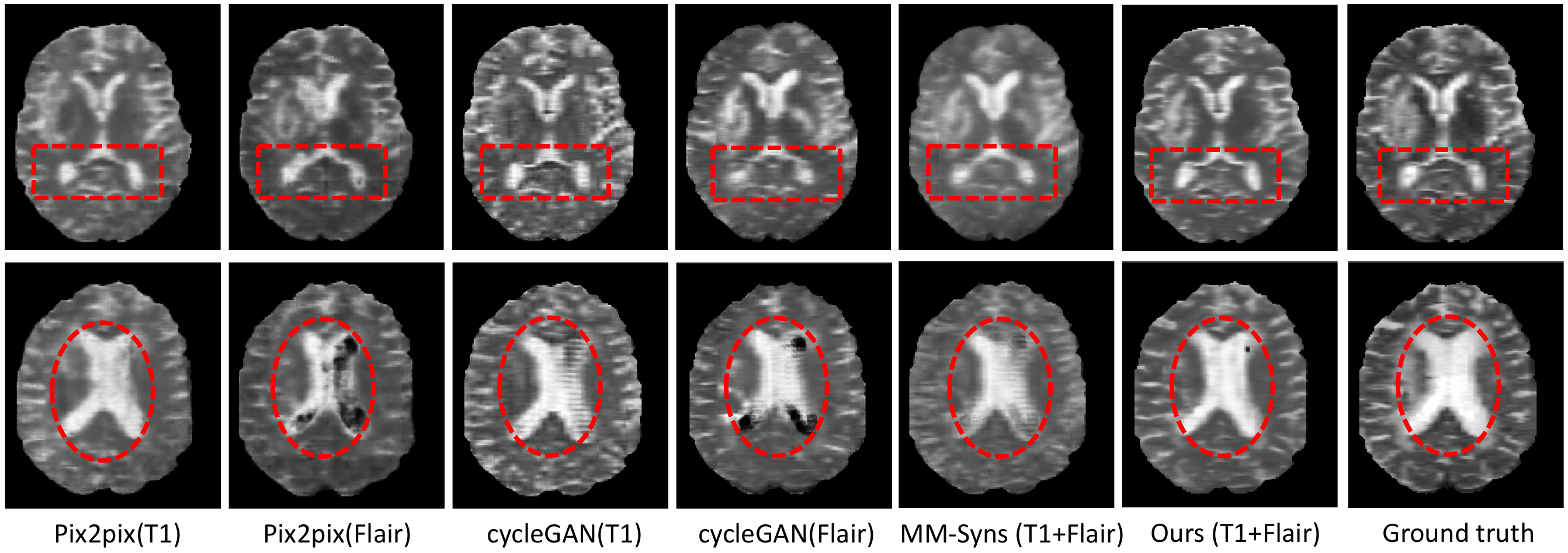}\vspace{-0.3cm}
			\caption{{Qualitative comparison between the proposed synthesis method and other state-of-the-art synthesis methods for the $T_2$ synthesis task on the ISLES2015 dataset. Note that our method and MM-Syns use both $T_1$ and $Flair$ modalities, while other methods use any one of the two}. }
			\label{fig61}
		\end{center}\vspace{-0.35cm}
	\end{figure*}

	In addition, we also evaluate the performance for synthesizing $T_2$ modality images using $T_1$ and $Flair$ images on the ischemic stroke lesion segmentation challenge 2015 (ISLES2015) dataset \cite{maier2017isles}. This dataset consists of multi-spectral MR images. In this study, we choose the sub-acute ischemic stroke lesion segmentation (SISS) cohort of patients. Each case consists of four sequences namely $T_1$, $T_2$, $DWI$ and $Flair$, and are rigidly co-registered to the $Flair$ sequence. More details about the preprocessing steps can be found in \cite{maier2017isles}. For a 2D slice ($230\times230$),  we crop out an image of size $160\times 180$ from the center region, and we also split each cropped image ($160\times180$) into four overlapping patches of size $128\times128$. Besides, we use 28 training cases and 17 testing cases in this study. For each volume, we also linearly scale the original intensity values to $[-1,1]$. Table~\ref{tab32} shows the quantitative evaluation results for this task. From Table~\ref{tab32}, as it can be seen that our method outperforms all comparison methods in all three metrics. Fig.~\ref{fig61} shows qualitative comparisons between the proposed synthesis method and other state-of-the-art methods in the $T_2$ synthesis task. As can be seen, our method achieves much better synthesis results. 

	\renewcommand\arraystretch{1.0}
	\begin{table}[t]
		\setlength{\belowdisplayskip}{0pt}
		\setlength{\abovedisplayskip}{0pt}
		\setlength{\abovecaptionskip}{0pt}
		\centering
		\scriptsize
		\caption{\footnotesize Quantitative evaluation results of the synthesized $T_2$ images using $T_1$ and $Flair$ or their combination for different methods on the ISLES2015 dataset (mean$\pm$standard deviation). Bold indicates the best result.}
		\begin{tabular}{p{3.2cm}|p{1.3cm}p{1.3cm}p{1.3cm}}  
			\toprule 
			\textbf{Methods}                                 & PSNR $\uparrow$    &NMSE $\downarrow	$        & SSIM $\uparrow$ \\
			\toprule
			Pix2pix ($T_1\rightarrow T_2$)                &21.70$\pm$0.906         &0.1009$\pm$0.012      &0.8228$\pm$0.036\\
			Pix2pix ($Flair\rightarrow T_2$)               &21.65$\pm$0.899         &0.1064$\pm$0.023      &0.8160$\pm$0.024\\
			cycleGAN ($T_1\rightarrow T_2$)            &20.12$\pm$0.609          &0.1502$\pm$0.021      &0.7734$\pm$0.028\\
			cycleGAN ($Flair\rightarrow T_2$)           &20.28$\pm$0.353         &0.1439$\pm$0.009      &0.7735$\pm$0.027\\
			MM-Syns ($T_1+Flair\rightarrow T_2$)    &20.36$\pm$0.704          &0.1419$\pm$0.017      &0.8223$\pm$0.025\\
			Ours  ($T_1+Flair\rightarrow T_2$)           &\textbf{22.46}$\pm$\textbf{0.854}      &\textbf{0.0881}$\pm$\textbf{0.015}      &\textbf{0.8414}$\pm$\textbf{0.030} \\
			
			\toprule
		\end{tabular}
		\label{tab32}
	\end{table}

	\subsection{Ablation Study}
	
\begin{figure}
		\begin{center}
			\includegraphics[width=0.4\textwidth]{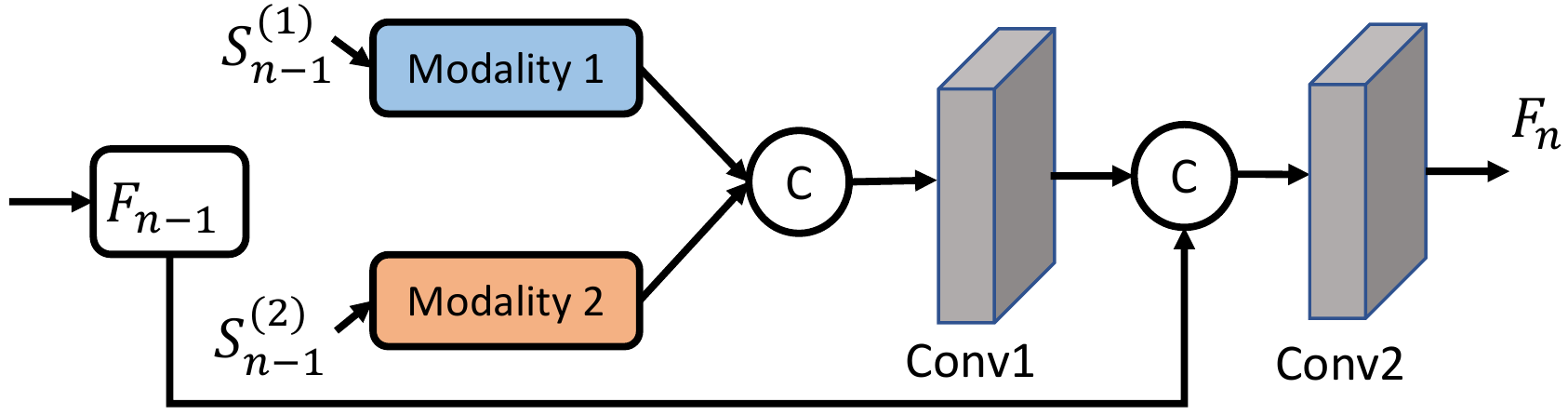}\vspace{-0.2cm}
			\caption{Illustration of the direct fusion strategy for multiple modalities.}
			\label{fig81}
		\end{center}
	\end{figure}

\begin{figure}
		\begin{center}
			\includegraphics[width=0.4\textwidth]{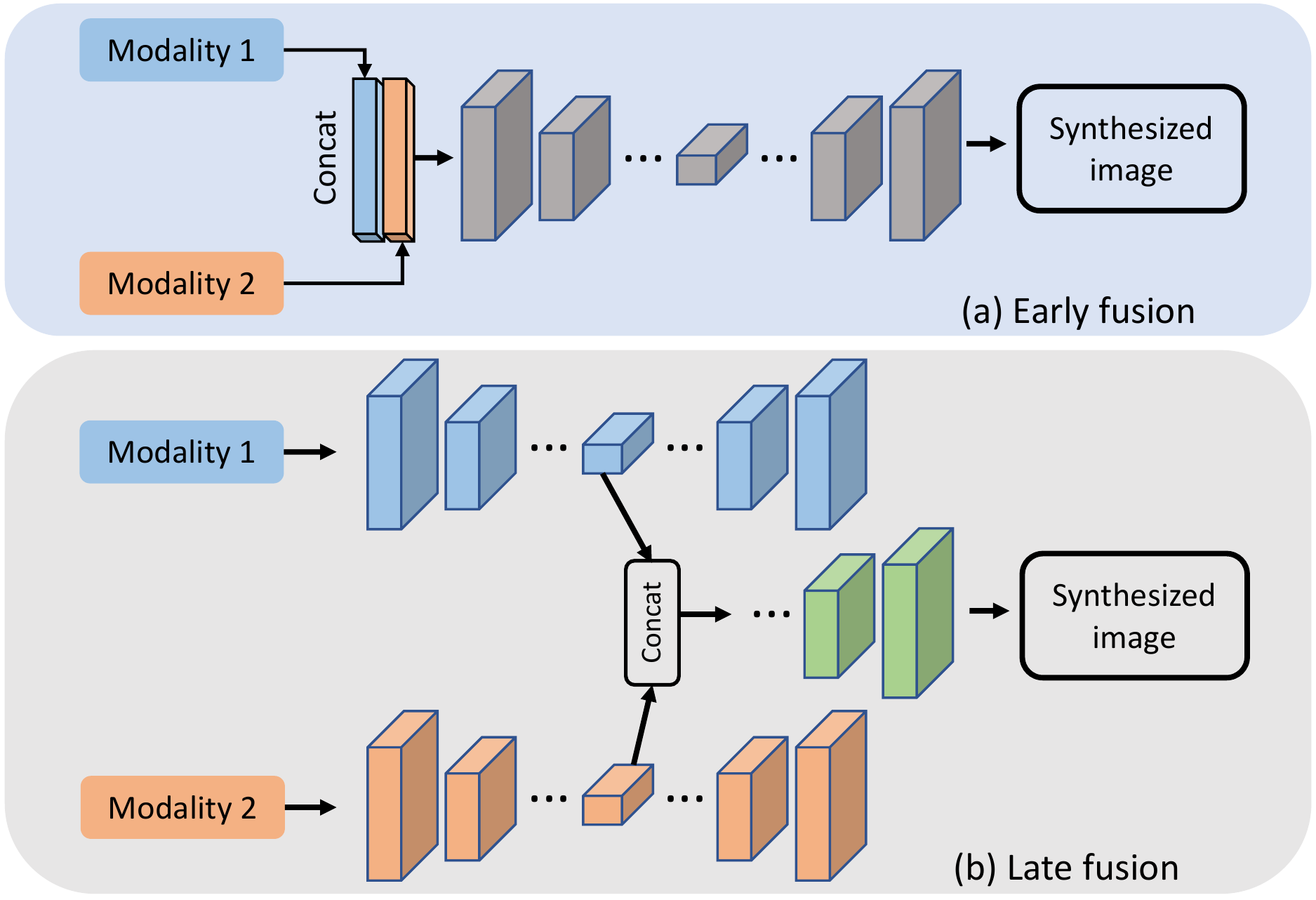}\vspace{-0.2cm}
			\caption{Illustration of two fusion strategies: (a) early fusion and (b) late fusion.}
			\label{fig8}
		\end{center}
	\end{figure}

	The proposed synthesis method (Hi-Net) consists of several key components, we conduct the following ablation studies to verify the importance of each one. \emph{First}, our model utilizes a novel MFB module to fuse multiple modalities or inputs. Conveniently, we define a direct concatenated strategy for fusing multiple modalities (denoted as ``ConcateFusion") as shown in Fig.~\ref{fig81}. To evaluate the effectiveness of the MFB module, we compare our full model with three degraded versions as follows: (1) We use the ``ConcateFusion" strategy both in the fusion network and generator network, denoted as ``Ours-degraded1"; (2) We use MFB modules in the fusion network and the ``ConcateFusion" in the generator network, denoted as ``Ours-degraded2"; and (3) We use the ``ConcateFusion" in the fusion network and MFB modules in the generator network, denoted as ``Ours-degraded3".  \emph{Second}, our model also employs a hybrid fusion strategy to fuse the various modalities. Thus, we compare the proposed Hi-Net with models using an early fusion strategy (as shown in Fig.~\ref{fig8}(a)) or a late fusion strategy (as shown in Fig.~\ref{fig8}(b)). We denote the above two strategies as ``Ours-earlyFusion" and ``Ours-lateFusion", respectively. Note that the skip connection components are included in both fusion strategies, even if they are not shown in Fig.~\ref{fig8}. 
	
	Table~\ref{tab04} shows quantitative evaluation results for the synthesized $Flair$ modality images using $T_1$ and $T_2$ when comparing our full model with its ablated versions. Compared with ``Ours-degraded1", it can be seen that our model with using the MFB module effectively improves the synthesis performance. This is because the proposed MFB module adaptively weights the different fusion strategies. Besides, when comparing ``Ours-degraded2" with ``Ours-degraded3", the results indicate that our model performs better when only using MFB modules in the fusion network than that only using MFB modules in the generator network. Further, compared with early and late fusions, the results demonstrate that our proposed hybrid fusion network performs better than both. This is mainly because our Hi-Net exploits the correlations among multiple modalities using the fusion network, and simultaneously preserves the modality-specific properties using the modality-specific network, resulting in better synthesis performance.

	\renewcommand\arraystretch{1.0}
	\begin{table}[t]
		\setlength{\belowdisplayskip}{0pt}
		\setlength{\abovedisplayskip}{0pt}
		\setlength{\abovecaptionskip}{0pt}
		\centering
		\scriptsize
		\caption{\footnotesize {Quantitative evaluation results of the synthesized $Flair$ images using $T_1$ and $T_2$ when comparing our full model with its ablated versions {on the BraTs2018 dataset (mean$\pm$standard deviation)}. Bold indicates the best result}.}
		\begin{tabular}{p{2.2cm}|p{1.3cm}p{1.3cm}p{1.3cm}}  
			\toprule 
			\textbf{Methods}           & PSNR $\uparrow$    &NMSE $\downarrow	$        & SSIM $\uparrow$ \\
			\toprule
			{Ours-degraded1}     &23.75$\pm$1.303         &0.0387$\pm$0.008      &0.8836$\pm$0.031 \\
			{Ours-degraded2}    &24.47$\pm$1.245          &0.0294$\pm$0.009      &0.8892$\pm$0.031 \\
			{Ours-degraded3}    &23.83$\pm$0.912          &0.0338$\pm$0.012      &0.8878$\pm$0.027\\
			{Ours-earlyFusion}   &22.84$\pm$0.975          &0.0421$\pm$0.009      &0.8612$\pm$0.032 \\
			Ours-lateFusion             &23.63$\pm$1.110           &0.0342$\pm$0.008      &0.8745$\pm$0.029\\
			Ours (Hi-Net)                 &\textbf{25.05}$\pm$\textbf{1.325}        &\textbf{0.0258}$\pm$\textbf{0.009}      &\textbf{0.8909}$\pm$\textbf{0.030}\\

			\toprule
		\end{tabular}
		\label{tab04}
	\end{table}

	\subsection{Discussion}
	
	In contrast to most existing methods, which focus on the single-input to single-output task \cite{Huang2017,Nie2017,Wolterink2017,Wang2018tmi,dar2019image,Yutmi2019}, our proposed model fuses multi-modal data (\emph{e. g.}, two modalities as input in our current study) to synthesize the missing modality images. It is worth noting that our method can effectively gather more information from the different modalities to improve the synthesis performance compared to using only a single modality as input. This is mainly because the multi-modal data can provide complementary information and exploit the properties of each modality. 
	
	Besides, our proposed Hi-Net consists of two modality-specific networks and one fusion network, where the modality-specific networks aim to preserve the modality-specific properties and the fusion network aims to exploit the correlations among multiple modalities. In the multi-view learning field, several studies focus on learning a common latent representation to exploit the correlations among multiple views \cite{zhou2020multi,zhang2018generalized}, while other methods explore complementary information. However, both of these are important for multi-view/modal learning \cite{zhou2019dual,gong2017exploring,zhou2015group}. For the proposed model, we have considered both aspects to improve fusion performance. 
	

	For our proposed model, one potential application is to first synthesize missing modality images and then use them to achieve a specific task. For example, in tumor segmentation and overall survival time prediction tasks, there exist different modality MR images, but it is common to have several missing modalities in clinical applications \cite{shen2019brain}. Thus, our framework can effectively synthesize the missing modality images and then perform multi-modal segmentation and overall survival prediction tasks using existing methods. Besides, it is widely known that a large amount of data is critical to success when training deep learning models. In practice, it is often difficult to collect enough training data, especially for a new imaging modality not well established in clinical practice yet. What's more, data with high-class imbalance or insufficient variability \cite{shin2018medical} often results in poor classification performance. Thus, our model can synthesize more multi-modal images, which can be regarded as supplementary training data to boost the generalization capability of current deep learning models.

	\section{Conclusion}
	\label{Conclusion}
	
	In this paper, we have proposed a novel end-to-end hybrid-fusion network for multi-modal MR image synthesis. Specifically, our method explores the modality-specific properties within each modality, and simultaneously exploits the correlations across multiple modalities. Besides, we have proposed a layer-wise fusion strategy to effectively fuse multiple modalities within different feature layers. Moreover, an MFB module is presented to adaptively weight different fusion strategies. The experimental results in multiple synthesis tasks have demonstrated that our proposed model outperforms other state-of-the-art synthesis methods in both quantitative and qualitative measures. In the future, we will validate whether the synthetic images as a form of data augmentation can boost the multi-modal learning performance.

\footnotesize
\bibliographystyle{IEEEbib}
\bibliography{tmi}

\end{document}